\documentclass[letterpaper, 10 pt, conference]{ieeeconf}

\makeatletter
\let\NAT@parse\undefined
\makeatother
\usepackage{amsmath}
\usepackage{amssymb}
\usepackage[numbers]{natbib}
\usepackage[dvipsnames]{xcolor}
\usepackage{hyperref}
\usepackage{booktabs}
\usepackage{graphicx}
\usepackage{siunitx}
\usepackage{tabularx}
\usepackage[export]{adjustbox}
\usepackage[capitalise]{cleveref}
\usepackage{tikz}
\usetikzlibrary{arrows.meta,positioning,calc,ext.node-families}
\usepackage{balance}

\crefname{equation}{eq.}{eqs.}

\IEEEoverridecommandlockouts
\graphicspath{ {./figures/} }

\makeatletter
\newcommand*{\@rowstyle}{}
\newcommand*{\rowstyle}[1]{%
  \gdef\@rowstyle{#1}%
  \@rowstyle\ignorespaces
}
\newcolumntype{=}{>{\gdef\@rowstyle{}}}
\newcolumntype{+}{>{\@rowstyle}}
\makeatother

\newcolumntype{Y}{>{\centering\arraybackslash}X}

\title{\LARGE \bf Particle Filtering with Generative Trajectory Models \\ for State Estimation and Control}

\author{Lukas C. Lao Beyer$^{1}$, Samvit Das$^{1}$ and Sertac Karaman$^{1}$%
\thanks{$^{1}$Laboratory for Information and Decision Systems, Massachusetts Institute of Technology, Cambridge, MA. {\tt\small \{llb,samvitd,sertac\}@mit.edu}}%
}

\begin{document}
\maketitle
\thispagestyle{empty}
\pagestyle{empty}

\begin{abstract}
    Generative models can support both state estimation and control in robotics.
For control, a common approach is to use the model as a policy that directly samples actions conditioned on observation histories.
However, models that jointly represent states, actions, and observations can also predict how the robot state and its environment evolve.
We investigate whether these predictions can improve state estimation and control through a change in inference procedure, without modifying the pretrained model.
To this end, we use particle filtering to maintain competing hypotheses about state histories.
A single pretrained diffusion model generates these hypotheses along with their predicted observations and associated action proposals.
As new measurements arrive, particles are reweighted according to how well their predicted observations match the measurements.
While particle filtering conventionally uses these weights only for state estimation, we also use them to guide action selection, favoring proposals associated with hypotheses that better explain the observations.
Simulation experiments in partially observed navigation and vision-based manipulation demonstrate improvements in both state estimation and task performance compared to action generation without filtering, using the same pretrained model and no additional test-time information.

\end{abstract}


\section{Introduction}

Diffusion- and flow-based generative models have become powerful tools for robot learning and control.
They have been used as policies that directly generate actions conditioned on observations, as world models that predict the evolution of the environment, and, more recently, as unified world-action models that jointly model states, observations, and actions over time.
Such joint trajectory models can support a variety of conditional generation tasks, depending on which variables are provided as context and which are generated.
Their usefulness therefore depends not only on the distributions they learn, but also on how those distributions are used at inference time.

For control, a natural approach is to directly generate actions conditioned on the available observation-action history.
In principle, if this procedure produces an optimal policy, no alternative inference strategy with access to the same information can achieve better expected task performance.
In practice, however, learned policies are generally imperfect due to limited training data, finite model capacity, and distribution shifts between training and deployment.
Moreover, a joint generative model may capture useful relationships between states, observations, and actions that are not fully exploited by direct action generation.
This creates an opportunity to improve control through additional inference-time computation, without modifying the learned model.
This question is particularly relevant under partial observability, where successful control depends on inferring the underlying state from observations accumulated over time.

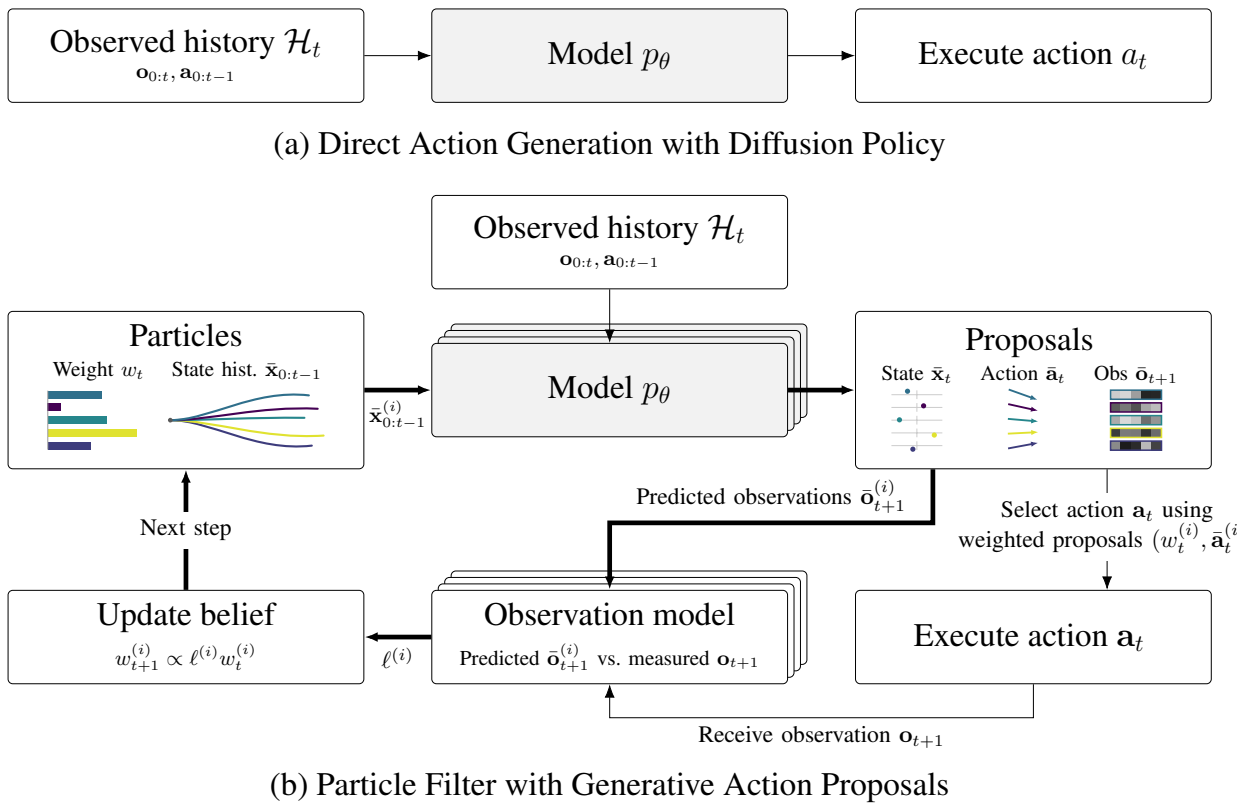
\begin{figure}[t]
  \centering
  \resizebox{\columnwidth}{!}{%
    \begingroup
\def\hgap{1cm}%

\NewDocumentCommand{\stacknode}{O{} O{} m m m}{%
  \node[#1,#2,opacity=0] (#3-position) {#5};%
  \pgfmathtruncatemacro{\last}{#4-1}%
  \foreach \i in {\last,...,0}{%
    \ifnum\i=0\relax
      \node[#1,anchor=center,alias=#3] (#3-\i)
        at ($(#3-position.center)+(0,0,-\i/4)$) {#5};
    \else
      \node[#1,anchor=center] (#3-\i)
        at ($(#3-position.center)+(0,0,-\i/4)$) {#5};
    \fi
  }%
}%

\definecolor{particleV0}{HTML}{440154}
\definecolor{particleV1}{HTML}{3B528B}
\definecolor{particleV2}{HTML}{21918C}
\definecolor{particleV3}{HTML}{5EC962}
\definecolor{particleV4}{HTML}{FDE725}

\newcommand{\setparticlecolor}[1]{%
  \pgfmathsetmacro{\t}{max(0,min(1,(#1-0.12)/(0.34-0.12)))}%
  \pgfmathtruncatemacro{\lo}{min(3,floor(4*\t))}%
  \pgfmathtruncatemacro{\hi}{\lo+1}%
  \pgfmathsetmacro{\mix}{100*(4*\t-\lo)}%
  \colorlet{particleColor}{particleV\hi!\mix!particleV\lo}%
}%

\def\particledata{%
  0.16/0.96/-0.38,%
  0.33/1.04/-0.23,%
  0.22/0.91/0.035,%
  0.05/1.00/0.17,%
  0.20/0.94/0.37%
}%

\newcommand{\particleweights}{%
  \begin{tikzpicture}[
      x=40mm,y=10mm,
      baseline=(current bounding box.center),
    ]
    \path[use as bounding box]
      (-0.02,-0.49) rectangle (0.40,0.49);

    \foreach \w/\endx/\endy [count=\i] in \particledata {
      \setparticlecolor{\w}
      \pgfmathsetmacro{\bary}{0.19*(\i-3)}
      \fill[particleColor,sharp corners]
        (0,{\bary-0.055}) rectangle (\w,{\bary+0.055});
    }

    \draw[black!40,line width=0.3pt]
      (0,-0.44) -- (0,0.44);
  \end{tikzpicture}%
}%

\newcommand{\particlehistories}{%
  \begin{tikzpicture}[
      x=22mm,y=10mm,
      baseline=(current bounding box.center),
    ]
    \path[use as bounding box]
      (-0.04,-0.55) rectangle (1.08,0.55);

    \foreach \w/\endx/\endy in \particledata {
      \setparticlecolor{\w}
      \draw[particleColor,line width=0.9pt,line cap=round]
        (0,0)
        .. controls ({0.28*\endx},0)
                and ({0.60*\endx},{1.25*\endy})
        .. (\endx,\endy);
    }

    \fill[black!65] (0,0) circle[radius=1pt];
  \end{tikzpicture}%
}%

\newcommand{\proposalcolumn}[1]{%
  \begin{tikzpicture}[
      x=10mm,y=10mm,
      baseline=(current bounding box.center),
    ]
    \path[use as bounding box]
      (0,-0.49) rectangle (1,0.49);

    \foreach \w/\endx/\endy [count=\i] in \particledata {
      \setparticlecolor{\w}
      \pgfmathsetmacro{\rowy}{0.19*(\i-3)}
      \pgfmathsetmacro{\statex}{0.2+4*(\endx-0.9)}

      \ifcase#1\relax
        \draw[black!20,line width=0.25pt]
          (0.12,\rowy) -- (0.88,\rowy)
          (0.5,{\rowy-0.065}) -- (0.5,{\rowy+0.065});
        \fill[particleColor]
          (\statex,{\rowy+0.14*\endy})
          circle[radius=1.1pt];

      \or
        \pgfmathsetmacro{\targetY}{-0.12}
        \pgfmathsetmacro{\angle}{atan2(\targetY-\rowy,1.4)}
        \draw[
            particleColor,line width=0.8pt,
            -{Latex[length=1.1mm,width=0.8mm]},
          ]
          ({0.5-0.23*cos(\angle)},
           {\rowy-0.23*sin(\angle)})
          --
          ({0.5+0.23*cos(\angle)},
           {\rowy+0.23*sin(\angle)});

      \or
        \foreach \j in {0,...,4} {
          \pgfmathtruncatemacro{\shade}{%
            15+mod(31*\i+23*\j+13*\j*\j+7*\i*\j,75)}

          \fill[black!\shade,sharp corners]
            ({0.12+0.152*\j},{\rowy-0.068})
            rectangle
            ({0.12+0.152*(\j+1)},{\rowy+0.068});
}%

\draw[particleColor,line width=0.65pt,sharp corners]
  (0.12,{\rowy-0.068})
  rectangle (0.88,{\rowy+0.068});
      \fi
    }
  \end{tikzpicture}%
}%

\begin{tikzpicture}[
    >=Latex,
    box/.style={
      draw,
      rounded corners=2pt,
      font=\Large,
      ext/node family/size=box,
      align=center,
      fill=white,
      inner sep=5pt,
    },
    pfbox/.style={
      draw,
      rounded corners=2pt,
      font=\Large,
      ext/node family/width=box,
      ext/node family/height=pfbox,
      align=center,
      fill=white,
      inner sep=5pt,
    },
    model/.style={
      box,
      fill=black!5,
    },
    input/.style={
      box,
    },
    output/.style={
      box,
    },
    pfinput/.style={
      pfbox,
    },
    pfoutput/.style={
      pfbox,
    },
    arrow/.style={
      ->,
      -{Latex[length=2mm, width=1.5mm]}
    },
    marrow/.style={
      arrow,
      line width=2pt,
    },
  ]

  \begin{scope}[
      name prefix=dp-,
      local bounding box=dp,
      anchor=south,
    ]
    \node[model] (model) {Model $p_\theta$};
    \node[input, left=\hgap of model] (hist) {
      Observed history $\mathcal{H}_t$\\[-2mm]
      \small $\mathbf{o}_{0:t}, \mathbf{a}_{0:t-1}$%
    };
    \node[output, right=\hgap of model] (act) {Execute action $a_t$};

    \draw[arrow] (hist) -- (model);
    \draw[arrow] (model) -- (act);

    \node[below=3mm of model] {\Large (a) Direct Action Generation with Diffusion Policy};
  \end{scope}

  \begin{scope}[
      name prefix=pf-,
      local bounding box=pf,
      anchor=north,
      shift={(dp-dp.south)},
      yshift=-2.6cm,
    ]
    \stacknode[model]{model}{4}{Model $p_\theta$}

    \node[input, above=8mm of model] (hist) {
      Observed history $\mathcal{H}_t$\\[-2mm]
      \small $\mathbf{o}_{0:t}, \mathbf{a}_{0:t-1}$%
    };

    \node[
      pfinput,
      left=\hgap of model,
    ] (particles) {
      \small
      \setlength{\tabcolsep}{2pt}
      \begin{tabular}{cc}
        \multicolumn{2}{c}{\Large Particles} \\[3pt]
        Weight $w_t$ & State hist. $\bar{\mathbf{x}}_{0:t-1}$ \\%
        \particleweights & \particlehistories \\%
      \end{tabular}
    };

    \node[
      pfoutput,
      right=\hgap of model,
    ] (samples) {
      \small
      \begin{tabular}{ccc}
        \multicolumn{3}{c}{\Large Proposals} \\[3pt]
        State $\bar{\mathbf{x}}_t$ & Action $\bar{\mathbf{a}}_t$ & Obs $\bar{\mathbf{o}}_{t+1}$ \\%
        \proposalcolumn{0} &
        \proposalcolumn{1} &
        \proposalcolumn{2} \\%
      \end{tabular}
    };

    \node[output, below=1.8cm of samples] (act) {
      Execute action $\mathbf{a}_t$
    };

    \stacknode[box][left=\hgap of act]{observe}{4} {
      Observation model\\
      \small Predicted $\bar{\mathbf{o}}^{(i)}_{t+1}$ vs. measured $\mathbf{o}_{t+1}$
    }

    \node[box, left=\hgap of observe] (update) {
      Update belief\\
      \small $w_{t+1}^{(i)} \propto \ell^{(i)} w_t^{(i)}$
    };

    \draw[marrow] (particles)
    -- node[midway, below]{\small $\bar{\mathbf{x}}_{0:t-1}^{(i)}$}
    (model);
    \draw[arrow] (hist) -- (model);
    \draw[marrow] (model) -- (samples);

    \draw[arrow] ([xshift=11mm]samples.south) --
    node[
      midway,
      yshift=0.55cm,
      fill=white,
      align=center
    ]{
      Select action $\mathbf{a}_t$ using \\
      weighted proposals $(w_t^{(i)}, \bar{\mathbf{a}}_t^{(i)})$
    }
    ([xshift=11mm]act.north);

    \draw[marrow] ([xshift=-15mm]samples.south)
    -- ++(0,-8mm)
    -| node[pos=0.25, above] {Predicted observations $\bar{\mathbf{o}}_{t+1}^{(i)}$}
    (observe.north);

    \draw[arrow] (act.south)
    -- ++(0,-0.5cm) coordinate (bend)
    -- (bend -| observe.south)
    node[midway, below] {Receive observation $\mathbf{o}_{t+1}$}
    -- (observe.south);

    \draw[marrow] (observe)
    -- node[midway, below]{$\ell^{(i)}$}
    (update);

    \draw[marrow] (update) --
    node[
      midway,
      yshift=0.3cm,
      fill=white,
      align=center
    ]{Next step}
    (particles);

    \node[below=1.2cm of observe] {\Large (b) Particle Filter with Generative Action Proposals};
  \end{scope}
\end{tikzpicture}
\endgroup%
  }%
  \vspace{-0.8em}
  \caption{
    Two inference procedures using the same pretrained model~$p_\theta$.
    (a)~Direct action generation.
    (b)~Particle filtering maintains weighted state histories and generates joint state-action-observation proposals.
    Weights guide action selection.
    Agreement between predicted and measured observations updates the belief for the next step.
    Stacked blocks denote per-particle computations.
  }
  \label{fig:method-overview}
\end{figure}

In this work, we explore particle filtering as an alternative inference-time approach to state estimation and control with pretrained world-action models (\cref{fig:method-overview}).
Under partial observability, several different state histories may be consistent with the observations available so far.
Maintaining a diverse set of plausible histories allows later observations to help resolve this ambiguity.
While traditional particle filters maintain a distribution over possible states of the environment, our approach also associates each state hypothesis with candidate future actions, generated jointly by the same model.
As a result, evidence about the underlying state can directly influence action selection, allowing particle filtering to serve both state estimation and control within a unified framework.
Importantly, this approach requires no additional filtering-specific training or separately learned dynamics, observation, or policy models.
We compare particle filtering against direct action generation using the same pretrained model.

We investigate the benefits and limitations of this approach through experiments in partially observed navigation and visuomotor robotic manipulation.
Navigation provides a controlled setting in which we systematically study the effects of training data, distribution shift, and different particle filtering configurations.
We further evaluate the approach on manipulation tasks with high-dimensional RGB observations and substantially different action spaces.
Our results demonstrate that inference-time particle filtering can improve state estimation and control without additional model training, while highlighting the conditions under which these improvements can be expected.

\section{Related Work}

\paragraph{Generative policies, trajectory models, and world-action models}

Diffusion models have emerged as expressive models for both robot policies and trajectory-level decision making.
Diffusion Policy represents multimodal visuomotor action distributions through conditional denoising~\citep{chi2023diffusionpolicy}, while Diffuser and Decision Diffuser formulate planning and control as conditional trajectory generation~\citep{janner2022diffuser,ajay2023conditional}.
More recently, world-action models jointly represent world evolution and executable robot actions within a unified generative model.
Unified World Models, for example, couples video and action diffusion such that a single model can serve as a policy, forward or inverse dynamics model, or observation generator~\citep{zhu2025unified}.
Diffusion Forcing provides a complementary mechanism for flexible conditioning over arbitrary subsets of sequential variables \cite{chen2024diffusionforcing}.
We build on these flexible conditioning capabilities by using a pretrained joint state-action-observation model as the primitive in an inference-time particle filtering procedure.

\paragraph{Particle filtering with learned models}

Prior learned particle-filter approaches replace hand-specified filtering models with learnable dynamics and observation components embedded in a differentiable particle-filter structure, allowing the resulting estimator to be optimized end-to-end~\citep{jonschkowski2018dpf,karkus2018pfnet}.
More recently, DiffPF uses a conditional diffusion model to generate filtering-posterior particles~\citep{wan2026diffpf}.
These methods use learned models to improve estimation, but their particles are used to represent state hypotheses rather than candidate decisions.
In our setting, a pretrained generative model jointly produces states, observations, and actions, so each particle simultaneously represents a state hypothesis and an associated control rollout.
Moreover, our generative model is trained independently of the filtering procedure and without filtering-specific training or learned components.

\paragraph{Particle beliefs for control}

Particle beliefs have also been used for decision making under partial observability.
Classical online POMDP methods represent uncertainty with particles and couple this belief representation to online planning, typically using repeated forward simulation and tree search to evaluate candidate actions~\citep{silver2010pomcp,sunberg2018online}.
These approaches demonstrate the value of particle-based beliefs for control, but generally require a separate test-time planning procedure on top of the filter.
In our setting, each particle instead carries an action rollout generated jointly with its state and observation hypotheses.
Observation-based particle weights can therefore be used directly for action selection, without a separate belief-space planning or search procedure.

\paragraph{Particle methods for diffusion inference}

Particle-based methods have also been used within diffusion inference to improve sampling or steer generation toward task-specific objectives.
Such approaches maintain and resample candidate diffusion samples according to guidance or reward potentials~\citep{du2025dynaguide,singhal2025general}.
This use of particles is orthogonal to ours: their particles exist within a single generative inference process, whereas our particles persist across environment timesteps as hypotheses over latent state and associated future trajectories, and are reweighted using newly observed sensor data.

\section{Preliminaries}\label{sec:preliminaries}

We consider filtering and control under partial observability, with state~$\mathbf{x}_t$, action~$\mathbf{a}_t$, observation~$\mathbf{o}_t$, and available history~$\mathcal H_t=(\mathbf{o}_{0:t},\mathbf{a}_{0:t-1})$.
Filtering estimates~$p(\mathbf{x}_{0:t}\mid\mathcal H_t)$.

\paragraph{Particle filtering}
A particle filter represents this posterior by weighted state trajectories,
\begin{equation}
p(\mathbf{x}_{0:t}\mid\mathcal H_t)
\approx\sum_{i=1}^{N}w_t^{(i)}\delta_{\mathbf{x}_{0:t}^{(i)}}(\mathbf{x}_{0:t}),
\end{equation}
with $\sum_i w_t^{(i)}=1$.
In a bootstrap filter~\citep{gordon1993novel}, each trajectory is extended using the dynamics model,
\begin{equation}
  \mathbf{x}_t^{(i)}
  \sim p_{\mathrm{dyn}}(
    \mathbf{x}_t\mid\mathbf{x}_{t-1}^{(i)},\mathbf{a}_{t-1}).
\end{equation}
The arriving observation reweights particles by their likelihood $p_{\mathrm{obs}}(\mathbf{o}_t\mid\mathbf{x}_t^{(i)})$, favoring hypotheses that better explain the measurement.

\paragraph{Diffusion forcing}
Diffusion models generate samples by starting from noise and iteratively denoising toward structured outputs.
Diffusion forcing~\citep{chen2024diffusionforcing} extends this procedure to sequences by sampling a separate noise level for each token during training.
At inference, known tokens can be held at zero or low noise levels, while queried tokens are denoised fully starting from pure noise.
This enables the same model to condition on different subsets of sequential variables and modalities~\citep{song2025history}.

\section{Method}
\label{sec:method}

We combine particle filtering with a learned trajectory diffusion model.
The central difference from a bootstrap particle filter is the proposal distribution: instead of advancing each particle with a one-step dynamics kernel, we use a learned generative model to propose future states, observations, and actions jointly.
The resulting particles are therefore both state-estimation hypotheses and candidate control rollouts.

\subsection{Flexible Conditional Trajectory Sampler}

Our method exposes the diffusion model through a flexible conditioning interface over state, action, and observation tokens.
Each model query fixes a subset of trajectory variables and samples the remaining variables jointly.
Let $q_x,q_a,q_o$ denote query index sets to be generated, and let $h_x,h_a,h_o$ denote conditioning index sets supplied to the model.
We write the resulting conditional sampler as
\begin{equation}
(\bar{\mathbf{x}}_{q_x},\bar{\mathbf{a}}_{q_a},\bar{\mathbf{o}}_{q_o})
\sim
p_\theta(\mathbf{x}_{q_x},\mathbf{a}_{q_a},\mathbf{o}_{q_o}
\mid
\mathbf{x}_{h_x},\mathbf{a}_{h_a},\mathbf{o}_{h_o}).
\label{eq:flexible-conditional-sampler}
\end{equation}
This construction uses the diffusion model as a flexible proposal distribution $p_\theta$.
By changing which variables are conditioned on and which are queried, the same model can act as a state-transition proposal, an observation predictor, a future-action generator, or a joint trajectory model.

\subsection{Particle Filter}

We now describe how this flexible sampling interface enables particle filtering for both state estimation and decision making at inference time.
Each iteration uses the new observation to reweight competing state-history hypotheses, then generates an action proposal conditioned on each history.
The updated weights determine how these proposals influence the executed action.

\subsubsection{Particle State}
At the current timestep $t$, each particle carries a hypothesized state history~$\bar{\mathbf{x}}_{0:t-1}^{(i)}$ and a corresponding weight~$w_{t-1}^{(i)}$.
In the first iteration, particle weights are initialized uniformly and the algorithm proceeds directly to action proposal.

\subsubsection{State and Observation Prediction}
Before incorporating the observation at the current time $t$, we predict the current state and observation for each particle, conditioned only on its state history hypothesis, past observations and actions~$\mathcal{H}_{t-1}=(\mathbf{o}_{0:t-1}, \mathbf{a}_{0:t-2})$, and the action commanded at the preceding step $\mathbf{a}_{t-1}$:
\begin{equation}
(\bar{\mathbf{x}}_{t}^{(i)}, \_, \bar{\mathbf{o}}_{t}^{(i)})
\sim
p_\theta(\cdot \mid \bar{\mathbf{x}}_{0:t-1}^{(i)}, \mathbf{a}_{t-1}, \mathcal{H}_{t-1}).
\end{equation}

\subsubsection{Observation-Consistent Weight Updates}
We update each particle's weight according to how well its prediction agrees with the new observation~$\mathbf{o}_t$, favoring histories that better explain the observed sensor stream.
Let~$\ell(\mathbf{o}_t;\bar{\mathbf{o}}_t^{(i)}, \bar{\mathbf{x}}_t^{(i)})$ denote this nonnegative observation score, which need not be normalized.
It may be an explicit sensor likelihood~$p_{\mathrm{obs}}(\mathbf{o}_{t}\mid\bar{\mathbf{x}}_{t}^{(i)})$ or another model-based compatibility score.
We may also avoid sensor-specific design by using an \emph{observation-consistency} score directly comparing the realized observation~$\mathbf{o}_t$ to the previously predicted observation~$\bar{\mathbf{o}}_{t}^{(i)}$.
Our implementation compares predicted and measured observations using cosine similarity in a learned latent observation space, as described in~\cref{sec:impl}.
The particle weights are then updated as
\begin{align}
\bar{w}_{t}^{(i)} &= w_{t-1}^{(i)}\ell(\mathbf{o}_{t};\bar{\mathbf{o}}_{t}^{(i)},\bar{\mathbf{x}}_{t}^{(i)}), \\
w_{t}^{(i)} &= \frac{\bar{w}_{t}^{(i)}}{\sum_{j=1}^N \bar{w}_{t}^{(j)}}.
\end{align}
Resampling is applied using the updated weights if the effective sample size becomes small.

\subsubsection{Action Proposal}

We sample current and future states and future actions for each particle, sharing the commanded action and observation history $\mathcal{H}_t=(\mathbf{o}_{0:t}, \mathbf{a}_{0:t-1})$ across particles:
\begin{equation}
  (\bar{\mathbf{x}}_{t:T}^{(i)}, \bar{\mathbf{a}}_{t:T-1}^{(i)}, \_)
  \sim p_\theta(
  \cdot
  \mid\bar{\mathbf{x}}_{0:t-1}^{(i)},\mathcal H_t
  ),
\end{equation}
where~$T - t$ is the planning horizon.
The newly sampled current state incorporates the current measurement~$\mathbf{o}_t$.
It replaces the prediction used for scoring and is appended to the particle's state history.

\subsubsection{Action Aggregation}

We use the particle weights to choose the next action from the proposals, evaluating two action-selection rules.
\emph{Best-particle action selection} executes the proposal associated with the highest-weight particle.
\emph{Weighted action averaging} instead combines the action proposals using the weighted average
\begin{equation}
  \mathbf{a}_t = \sum_{i=1}^N w_t^{(i)}\bar{\mathbf{a}}_t^{(i)}.
\end{equation}
The selected action is then executed and the algorithm proceeds to the next iteration.

\section{When Can Particle Filtering Help?}
\label{sec:why-pf}

Particle filtering and direct action generation both operate using the same observation-action history.
If direct action generation already realizes an optimal policy with exact sampling, filtering cannot improve expected task performance.
Its potential benefit lies in exploiting learned relationships among states, observations, and actions that an imperfect direct policy may underuse.
Maintaining and reweighting explicit state hypotheses could expose this structure to action selection.
We hypothesize that this improves control when three conditions hold:

\begin{enumerate}
\item\label{cond:state-action-alignment}
  \textbf{State-action alignment.}
  The generative model must capture meaningful relationships between states and actions, such that more accurate or task-relevant state hypotheses tend to produce higher-quality action proposals.

\item\label{cond:obs-state-alignment}
  \textbf{Observation-state alignment.}
  The generative model must capture meaningful relationships between states and observations, such that compatibility between predicted and realized observations provides useful evidence about the quality of the corresponding state hypotheses.

\item\label{cond:coverage}
  \textbf{Sufficient particle coverage and diversity.}
  The generative model must assign sufficient probability to high-quality state-action hypotheses, and the particle population must retain enough diversity to represent these hypotheses until observations can distinguish between them.
\end{enumerate}

These conditions allow observation-based particle weighting to favor state hypotheses associated with better control proposals.
In \cref{sec:experiments}, we test whether each of the conditions holds in practice through a series of experiments.

\newcommand{\condref}[1]{Condition~\ref{#1}}

\section{Implementation}
\label{sec:impl}

\subsection{Latent Observation World-Action Model}

Raw sensor observations, such as RGB images, are high-dimensional and may contain substantial task-irrelevant variation.
This can make defining an observation-consistency metric $\ell$ directly in the raw observation space challenging.
We therefore jointly learn an observation encoder
$\mathcal{E}_{\mathrm{obs}}$ and a diffusion model over states, actions, and latent observations,
\begin{equation}
    \mathbf{o}_t
    =
    \mathcal{E}_{\mathrm{obs}}
    \left(\mathbf{o}^{\mathrm{raw}}_t\right).
\end{equation}
The resulting latent observations provide a compact representation for both trajectory modeling and the observation-consistency score $\ell$.

Directly backpropagating the latent-observation prediction loss into
the parameters of $\mathcal{E}_{\mathrm{obs}}$ may encourage a degenerate representation that is easy to predict but uninformative. We therefore train the encoder only through the prediction of task variables. Let
\begin{equation}
  \widetilde{\mathbf{x}} \sim q_{\tau_x}\!(\cdot \mid \mathbf{x}), \quad
  \widetilde{\mathbf{a}} \sim q_{\tau_a}\!(\cdot \mid \mathbf{a}), \quad\text{and}\quad
  \widetilde{\mathbf{o}} \sim q_{\tau_o}\!(\cdot \mid \mathbf{o})
\end{equation}
denote states, actions and latent observations~${\mathbf{o} = \mathcal{E}_{\mathrm{obs}}\left(\mathbf{o}^{\mathrm{raw}}\right)}$ corrupted by the diffusion forward process $q_\tau$ using diffusion noise levels $\tau$ which are sampled independently for each timestep and each input modality.
Let $f_{\theta}$ denote the denoising network, conditioned on the diffusion noise level of each token (we suppress this conditioning below for readability) and compute
\begin{equation}
    \widehat{\mathbf{x}},\widehat{\mathbf{a}},\_
    =
    f\!
    \left(
        \widetilde{\mathbf{x}},
        \widetilde{\mathbf{a}},
        \widetilde{\mathbf{o}}
    \right),
\end{equation}
where gradients are allowed to flow through
$\widetilde{\mathbf{o}}$ to $\mathcal{E}_{\mathrm{obs}}$.
We then compute the observation prediction using detached latent observations,
\begin{equation}
    \_,\_,\widehat{\mathbf{o}}
    =
    f\!
    \left(
        \widetilde{\mathbf{x}},
        \widetilde{\mathbf{a}},
        \operatorname{sg}(\widetilde{\mathbf{o}})
    \right),
\end{equation}
where $\operatorname{sg}$ denotes the stop-gradient operator.
Both passes share model parameters and use the same corrupted training example, differing only in the gradient flow to the observation encoder.

Using the clean-data-prediction parameterization for notational simplicity, the training objective is
\begin{equation}
    \mathcal{L}
    =
    \left\|
        \widehat{\mathbf{x}}-\mathbf{x}
    \right\|_2^2
    +
    \left\|
        \widehat{\mathbf{a}}-\mathbf{a}
    \right\|_2^2
    +
    \lambda_o
    \left\|
        \widehat{\mathbf{o}}
        -
        \operatorname{sg}(\mathbf{o})
    \right\|_2^2.
\end{equation}
Thus, the observation encoder is optimized to retain information useful for predicting task variables, while the diffusion model is simultaneously trained to model the resulting latent observation sequence.

\subsection{Latent Cosine Similarity Observation-Consistency Metric}

Our choice of observation-consistency metric used to compare predicted and realized observations in the particle filter uses cosine similarity of latent observations:
\begin{equation}
  \ell_{\cos}(\mathbf{o}_t; \bar{\mathbf{o}}^{(i)}_t)
  = \exp\!\left[
    \frac{\mathbf{o}_t^\top \bar{\mathbf{o}}^{(i)}_t}{
      \gamma
      \|\mathbf{o}_t\|
      \|\bar{\mathbf{o}}_t^{(i)}\|
    }
    \right],
  \label{eq:cos-obs-model}
\end{equation}
where $\gamma$ is a temperature scaling hyperparameter.

\section{Experiments}
\label{sec:experiments}

\newcommand{\gtrow}{\rowstyle{\color{black!45}}}
\newcommand{\maprow}{\rowstyle{\color{black!75}}}
\definecolor{ciGreen}{HTML}{16803C}
\definecolor{ciRed}{HTML}{B42318}
\definecolor{ciYellow}{HTML}{986700}
\newcommand{\pairedCI}[3]{{\color{#1}\scriptstyle[#2,#3]}}
\newcommand{\pairedCIs}[3]{{\color{#1}\scriptscriptstyle[#2,#3]}}

We evaluate our approach in two domains: goal-directed navigation in partially observed 2D environments and visuomotor robotic manipulation.
We primarily use navigation as a controlled testbed, where synthetic data generation and access to ground-truth state enable systematic studies of model quality, observation consistency, and particle filtering behavior.
We then evaluate our method on robotic manipulation tasks with high-dimensional RGB observations and substantially different action spaces, demonstrating its applicability beyond the navigation setting.

For each task and training-dataset size, all inference variants use the same trained trajectory model and observation encoder, with identical parameters.
The baseline denoted \emph{Diffusion Policy} performs direct action generation using this model without particle filtering.
At inference, conditioning histories are restricted to a rolling window that fits the model's context length.
Navigation executes one action per update, while manipulation executes eight actions open-loop.
For chunked execution, state and observation prediction conditions on the full previously commanded action chunk, and the observation scores over that chunk are multiplied in a single weight update.

\subsection{Experimental Setup (Navigation)}

Our navigation task requires collision-free traversal of an unknown 2D environment towards a goal specified in the frame of the robot's initial pose.
At each step, the robot receives a 2D LiDAR observation consisting of $N_{\rm rays}$ distance measurements and executes a~$[\Delta x, \Delta y] \in \mathbb{R}^2$ control action, specified in the robot's body frame.
To ensure that LiDAR-based localization becomes essential, we add zero-mean Gaussian noise to executed actions with a standard deviation of $40\%$ of the commanded step length.
Heading $\theta$ is not controlled by the policy, instead evolving according to an exogenous random walk process.
At each timestep, the state is the 4D vector~$[g_x^{\rm body}, g_y^{\rm body}, \sin(\Delta\theta), \cos(\Delta\theta)]$, combining the goal's position in the agent's current body frame with its heading relative to the trajectory's initial pose.
This representation is invariant to global translation and rotation, while the sine-cosine encoding avoids angular discontinuities.

\subsubsection{Dataset}

We generate a synthetic dataset containing~\num[group-separator={,}]{1000} examples with randomly placed circular and rectangular obstacles and randomly selected start and goal locations.
Each demonstration trajectory is created by first planning the shortest collision-free path between the start and goal.
Then, the pure-pursuit controller~\citep{coulter1992implementation} is used to simulate tracking of the planned reference under noisy action execution, and heading is assigned according to a correlated random walk.

\subsubsection{Model \& Training}

We implement the observation encoder~$\mathcal{E}_{\rm obs}$ using circular convolutions, producing one or more latent observation tokens per timestep.
The denoising network~$f$ is implemented as a diffusion transformer~\citep{peebles2023scalable} which receives noised latent observation, state, and action tokens.
A block-triangular attention mask is used to implement temporal causal masking.
Goal conditioning is achieved through in-context conditioning in which the first timestep's state is set to the relative goal position.
Since the goal is provided only through the initial state, the navigation Diffusion Policy baseline retains its sampled state history to carry goal information forward as the context window advances.

\subsubsection{Evaluation Metrics and Protocol}

We measure task-relevant performance using the deviation between the goal position and the position at termination.
State-estimation metrics use weighted mean particle positions,
independently of action selection.
We report ADE (mean absolute position error) and translational RPE
(mean relative-translation error over consecutive intervals of distance $0.1$).
Unless otherwise stated, navigation experiments use 200 held-out evaluation scenarios.
Each experiment is repeated eight times using different random seeds.
Tables report mean metrics and paired 95\% bootstrap confidence intervals for percentage changes in mean performance relative to Diffusion Policy.
We average repeated runs within each scenario and resample scenarios, using the same sampled scenarios for each method and its baseline.
When applicable,~32 candidates or particles are used, unless noted otherwise.
Unless otherwise stated, navigation particle-filter experiments use best-particle action selection.

\subsection{Comparison To Other Candidate-Selection Methods}

\begin{table}[t]
\centering
\caption{
  Full particle filtering outperforms na\"ive action selection under the pose-oracle observation model.
}
\label{tab:oracle-pf-ablation}
\vspace{-0.5em}
\small
\setlength{\tabcolsep}{3pt}
\resizebox{\columnwidth}{!}{%
\begin{tabular}{lcc c@{\hspace{1pt}}cc@{\hspace{1pt}}cc@{\hspace{1pt}}c}
\toprule
Method
& \shortstack{Retained\\histories}
& \shortstack{Accum.\\weights}
& \multicolumn{2}{c}{Final dist $\downarrow$}
& \multicolumn{2}{c}{ADE $\downarrow$}
& \multicolumn{2}{c}{RPE $\downarrow$} \\
\midrule

Diffusion Policy
& 1 & --
& $0.898$ &
& $0.494$ &
& $0.116$ & \\

One-step selection
& 1 & No
& $0.528$ & $\pairedCIs{ciGreen}{-44\%}{-38\%}$
& $0.253$ & $\pairedCIs{ciGreen}{-51\%}{-47\%}$
& $0.067$ & $\pairedCIs{ciGreen}{-43\%}{-40\%}$ \\

Multi-history sel.
& 32 & No
& $0.589$ & $\pairedCIs{ciGreen}{-37\%}{-32\%}$
& $0.282$ & $\pairedCIs{ciGreen}{-46\%}{-40\%}$
& $0.092$ & $\pairedCIs{ciGreen}{-24\%}{-15\%}$ \\

\textbf{Full PF}
& 32 & Yes
& $\mathbf{0.426}$ & $\pairedCIs{ciGreen}{-56\%}{-49\%}$
& $\mathbf{0.195}$ & $\pairedCIs{ciGreen}{-62\%}{-59\%}$
& $\mathbf{0.061}$ & $\pairedCIs{ciGreen}{-49\%}{-45\%}$ \\

\bottomrule
\end{tabular}%
}
\end{table}

We begin by investigating how particle filtering compares to other methods that use selection from among several candidates.

\subsubsection{Ground-truth pose oracle likelihood}

To evaluate state-action alignment independently of observation model accuracy, we first use a privileged pose oracle score that assigns higher likelihood to particles whose predicted poses agree more closely with ground truth:
\begin{equation}
  \ell_{\mathrm{oracle}}^{(i)}(\mathbf{x}_t;\mathbf{\bar x}_t^{(i)})
  = \exp\!\left[
    -\Delta\bar{\mathbf p}_t^{(i)} / \gamma_p
    -\Delta\bar\theta_t^{(i)} / \gamma_\theta
  \right].
\end{equation}
Here, $\Delta\bar{\mathbf p}_t^{(i)} = \|\bar{\mathbf p}_t^{(i)}-\mathbf p_t\|$ refers to the position error (positions $\mathbf{p}$ share the initial-pose frame), and each particle's heading error $\Delta\bar\theta_t^{(i)}$ is defined analogously.
The scales~$\gamma_p$ and~$\gamma_\theta$ are hyperparameters.

\subsubsection{Baselines}

Our first baseline, \emph{one-step selection}, maintains a single state trajectory history.
At each step, it samples and executes an action conditioned on that history and the latest observation.
It then compares $N$ candidate extensions of the history, selecting the one that most accurately predicts the resulting ground-truth pose. The selected history is retained for the next step.

\emph{Multi-history selection} instead maintains $N$ distinct state histories.
At each update, it selects the action proposal associated with the highest current pose-oracle score.
Compared to full particle filtering, selection uses only the latest prediction error, without accumulating scores over time.

\subsubsection{Results (\texorpdfstring{\cref{tab:oracle-pf-ablation}}{Table \ref*{tab:oracle-pf-ablation}})}

We find that both baselines are able to leverage ground-truth pose-informed state selection to achieve improved task performance and state estimation accuracy compared to diffusion policy, indicating that the model has indeed learned significant state-action alignment, providing evidence for \condref{cond:state-action-alignment}.
However, simple one-step selection that does not persist diverse histories and multi-history selection that does not accumulate weights across timesteps both perform significantly worse than full particle filtering.

\subsection{Observation-Consistency Observation Model}

\begin{figure*}[t]
  \centering
  \small
  \begin{tabularx}{\textwidth}{>{\centering\arraybackslash}m{2.2cm}@{}X}
    & {%
        \begin{tabularx}{1.011\linewidth}{lYYYYY}
        \hspace{-0.1cm} & $t=2$ & $t=8$ & $t=20$ & $t=30$ & $t=41$ \\
        \end{tabularx}%
      }\\
    Diffusion Policy &     \includegraphics[valign=c,width=\linewidth]{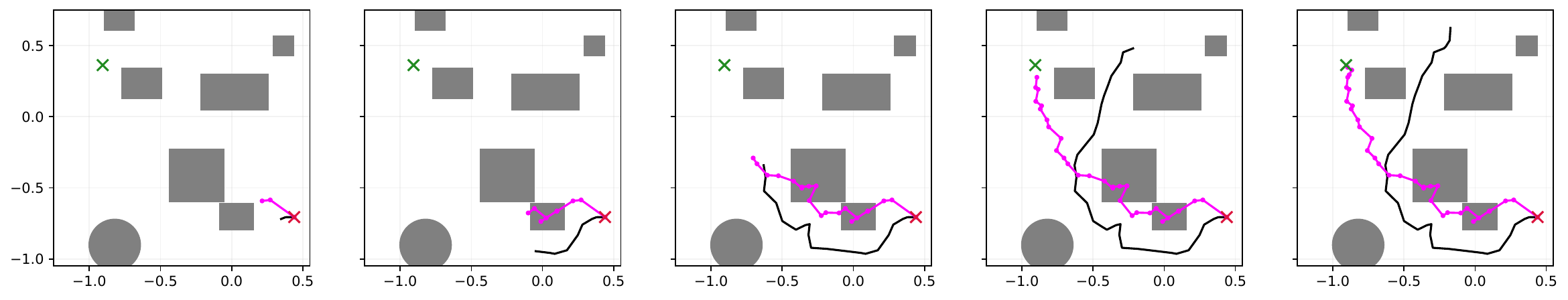} \\[-0.25cm]
    Particle Filter &     \includegraphics[valign=c,width=\linewidth]{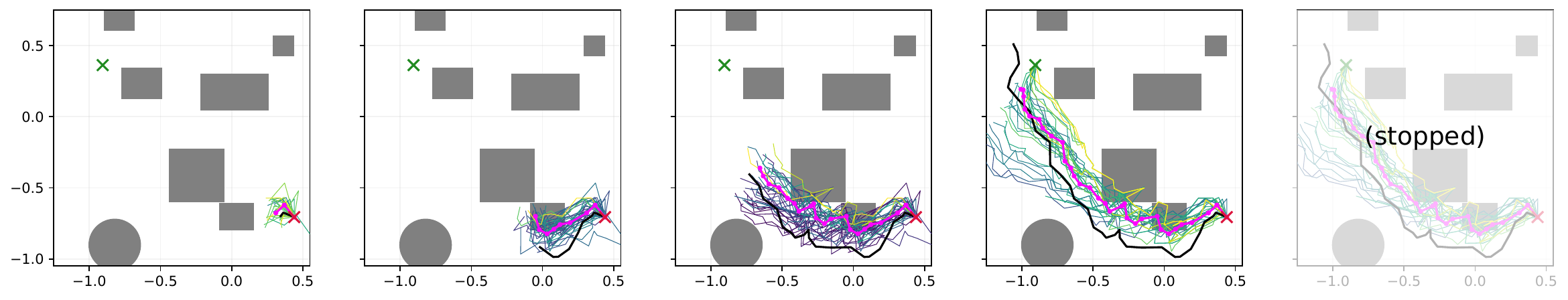} \\
  \end{tabularx}
  \caption{
    Closed-loop navigation with Diffusion Policy (top) and particle filtering (bottom).
    Magenta: estimated path; black: executed path; green/red crosses: goal/start.
    Particle brightness indicates weight.
    Diffusion Policy stops at $t=41$, particle filter stops at $t=30$.
  }
  \label{fig:pf-planner-example}
  \vspace{-0.3cm}
\end{figure*}

\begin{table}[t]
\centering
\caption{
  Observation-consistency observation model enables improved task performance and absolute state estimation.
}
\label{tab:pf-nav-comparison}
\vspace{-0.5em}
\setlength{\tabcolsep}{2pt}
\resizebox{\columnwidth}{!}{%
\begin{tabular}{=l +c@{\hspace{1pt}} +c +c@{\hspace{1pt}} +c +c@{\hspace{1pt}} +c}
\toprule
\hspace{0.5em} Method
& \multicolumn{2}{c}{Final dist $\downarrow$}
& \multicolumn{2}{c}{ADE $\downarrow$}
& \multicolumn{2}{c}{RPE $\downarrow$} \\
\midrule

\emph{Baselines:}&&&&&&\\

\hspace{0.5em} Diffusion Policy
& $0.898$ &
& $0.4941$ &
& $0.1155$ & \\

\hspace{0.5em} Ensemble
& $0.940$ & $\pairedCI{ciYellow}{-1\%}{+10\%}$
& $0.4516$ & $\pairedCI{ciGreen}{-16\%}{-2\%}$
& $0.0658$ & $\pairedCI{ciGreen}{-44\%}{-40\%}$ \\

\hspace{0.5em} PF w/ shuffled weights
& $0.874$ & $\pairedCI{ciYellow}{-6\%}{+1\%}$
& $0.4383$ & $\pairedCI{ciGreen}{-15\%}{-8\%}$
& $0.0703$ & $\pairedCI{ciGreen}{-40\%}{-37\%}$ \\

\hspace{0.5em} PF (weighted mean)
& $0.885$ & $\pairedCI{ciYellow}{-7\%}{+4\%}$
& $0.4196$ & $\pairedCI{ciGreen}{-22\%}{-9\%}$
& $\mathbf{0.0653}$ & $\pairedCI{ciGreen}{-44\%}{-41\%}$ \\

\midrule

\multicolumn{7}{l}{\emph{With privileged information:}} \\

\maprow \hspace{0.5em} PF w/ known map
& $0.633$ & $\pairedCI{ciGreen}{-33\%}{-26\%}$
& $0.3143$ & $\pairedCI{ciGreen}{-40\%}{-33\%}$
& $0.0677$ & $\pairedCI{ciGreen}{-43\%}{-38\%}$ \\

\gtrow \hspace{0.5em} PF w/ pose oracle
& $0.426$ & $\pairedCI{ciGreen}{-56\%}{-49\%}$
& $0.1947$ & $\pairedCI{ciGreen}{-62\%}{-59\%}$
& $0.0609$ & $\pairedCI{ciGreen}{-49\%}{-45\%}$ \\

\midrule

\textbf{Particle Filter}
& $\mathbf{0.812}$ & $\pairedCI{ciGreen}{-13\%}{-6\%}$
& $\mathbf{0.4087}$ & $\pairedCI{ciGreen}{-20\%}{-14\%}$
& $0.0697$ & $\pairedCI{ciGreen}{-41\%}{-37\%}$ \\

\bottomrule
\end{tabular}%
}
\end{table}

In the previous section, we have demonstrated that particle filtering provides a principled way of incorporating scores assigned to multiple candidates.
However, in practice, access to an oracle may not be available.
We therefore investigate the effectiveness of the proposed cosine similarity observation-consistency score~(eq.~\ref{eq:cos-obs-model}) as an observation model.

To distinguish the benefits of particle filtering from those of generating and combining multiple proposals, we evaluate two additional baselines.
The ensemble baseline uses the same pretrained model to generate $N$ proposals and uniformly averages their state estimates and proposed actions, providing a comparison with similar proposal-generation cost.
Each ensemble member retains its own sampled state history across planning iterations.
Particle filtering with shuffled weights randomly permutes the observation-model scores among particles at each update, before applying them to the accumulated weights.
This disrupts the correspondence between each particle hypothesis and its observation-consistency score.

\subsubsection{Known-map observation model}
To establish that cosine-similarity observation-consistency is a suitable score, we must first verify that the latent observation representation learned by the model supports meaningful comparison via cosine similarity.
To this end, we consider a \emph{known-map} observation model
\begin{equation}
  \ell_{\rm map}(\mathbf{o}_t; \mathbf{\bar x}_t^{(i)})
  = \ell_{\cos}(\mathbf{o}_t; \mathcal{E}_{\rm obs}(
  \mathtt{raycast}(\mathbf{\bar x}_t^{(i)})
  )),
  \label{eq:known-map-obs-model}
\end{equation}
where $\mathtt{raycast}(\mathbf{x})$ denotes the result of computing a LiDAR observation by raycasting from the pose $\mathbf{x}$ in the known map.
As shown in \cref{tab:pf-nav-comparison}, this known-map observation model enables a substantial improvement across both task performance and state estimation metrics, demonstrating that the latent observation representation is indeed suitable for use with the proposed cosine-similarity-based observation model.

\subsubsection{Predicted observation consistency}
We further verify whether the model has learned to predict latent observations that are consistent and responsive to each particle's state history.
As shown in \cref{tab:pf-nav-comparison}, using the cosine observation model (eq.~\ref{eq:cos-obs-model}) to compare each particle's predicted latent observation with the latent representation of the measured observation significantly outperforms the Diffusion Policy baseline, supporting \condref{cond:obs-state-alignment}.
The ensemble and shuffled-weight baselines also improve state estimation, but neither yields a clear improvement in task performance.
In particular, the comparison with shuffled scores indicates that observation-consistent weighting helps identify state-action hypotheses that lead to better control.
In \cref{fig:pf-planner-example}, we show an example of particle filtering using the predicted-observation consistency model, alongside the Diffusion Policy baseline.

\subsection{Action Selection Rule}
\Cref{tab:pf-nav-comparison} additionally compares best-particle action selection with weighted action averaging, using the same particle-filtering procedure and observation-consistency score.
Weighted action averaging achieves similar ADE and lower RPE than best-particle selection, yet only best-particle selection yields a clear improvement in final goal distance over Diffusion Policy.
These results suggest that particle filtering’s estimation benefits are robust to the choice between these two rules, while using the resulting belief effectively for control requires an appropriate action-selection rule.

\subsection{Particle Count Scaling}

\begin{figure}
  \centering
  \includegraphics[width=\columnwidth]{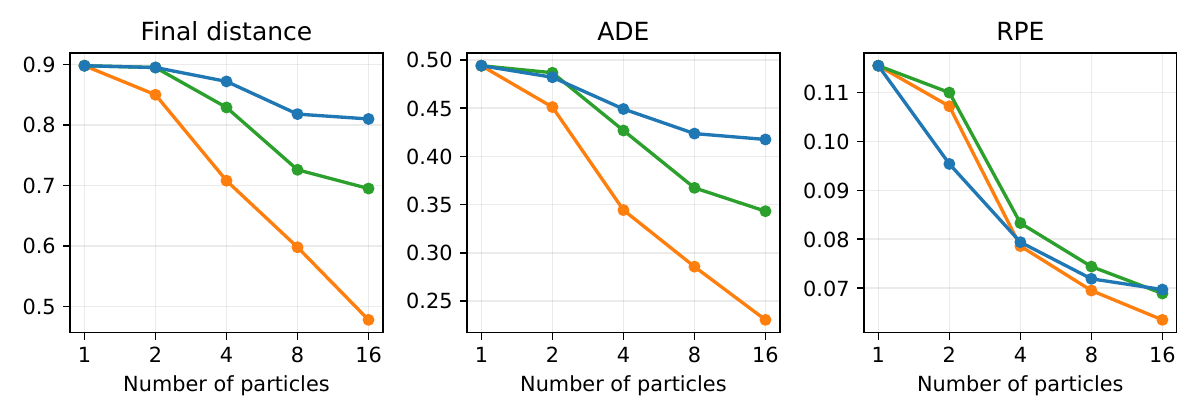}%
  \vspace{-0.8em}
  \caption{Particle-count scaling with pose-oracle (\textcolor{BurntOrange}{orange}), known-map (\textcolor{ForestGreen}{green}), and predicted-observation (\textcolor{NavyBlue}{blue}) weighting.}
  \label{fig:num-particle-sweep}
\end{figure}

Increasing the particle count tests whether the model supplies sufficient hypothesis coverage (\condref{cond:coverage}).
\Cref{fig:num-particle-sweep} shows improvements in task performance and state estimation under all three observation models.
Predicted-observation weighting exhibits diminishing returns beyond 16 particles, while the pose oracle continues to improve, supporting both coverage and state-action alignment.
Known-map weighting achieves intermediate task performance and ADE.
The remaining gap to the pose oracle may reflect both ambiguity in the observations and limitations of the learned representation and cosine similarity score.

\subsection{Dataset Scaling}

\begin{table}[t]
\centering
\caption{
  Final distance achieved by navigation models trained on datasets of increasing size.
}
\label{tab:dataset-size-comparison}
\vspace{-0.5em}
\setlength{\tabcolsep}{4pt}
\resizebox{\columnwidth}{!}{%
\begin{tabular}{=l +c@{\hspace{1pt}} +c +c@{\hspace{1pt}} +c +c@{\hspace{1pt}} +c}
\toprule
& \multicolumn{6}{c}{Training Dataset Size} \\
\cmidrule(lr){2-7}
Method
& \multicolumn{2}{c}{1,000}
& \multicolumn{2}{c}{10,000}
& \multicolumn{2}{c}{100,000} \\
\midrule

Diffusion Policy
& $0.898$ &
& $0.454$ &
& $\mathbf{0.369}$ & \\

\textbf{Particle Filter}
& $\mathbf{0.812}$ & $\pairedCI{ciGreen}{-13\%}{-6\%}$
& $\mathbf{0.415}$ & $\pairedCI{ciGreen}{-13\%}{-4\%}$
& $0.395$ & $\pairedCI{ciRed}{+4\%}{+11\%}$ \\

\gtrow PF w/ GT oracle
& $0.426$ & $\pairedCI{ciGreen}{-56\%}{-49\%}$
& $0.237$ & $\pairedCI{ciGreen}{-52\%}{-44\%}$
& $0.231$ & $\pairedCI{ciGreen}{-42\%}{-33\%}$ \\

\bottomrule
\end{tabular}%
}
\end{table}

We view particle filtering as a way to exploit learned state-action-observation relationships that direct action generation may not fully utilize, particularly when training data is limited.
We therefore test increasing training data from 1,000 to 100,000 demonstrations, which substantially improves direct action generation (\cref{tab:dataset-size-comparison}).
Observation-consistency filtering helps at 1,000 and 10,000 demonstrations but performs slightly worse at 100,000.
Pose-oracle filtering retains a substantial advantage throughout, indicating that the model continues to produce state-action hypotheses supporting better control, and showing that particle filtering can remain an effective way of incorporating additional known information.

\begin{table}[t]
\centering
\caption{
  On OOD scenarios, particle filtering helps even using the model trained on 100,000 scenarios.
}
\label{tab:ood-100k-comparison}
\vspace{-0.5em}
\setlength{\tabcolsep}{2pt}
\resizebox{\columnwidth}{!}{%
\begin{tabular}{=l +c@{\hspace{1pt}} +c +c@{\hspace{1pt}} +c +c@{\hspace{1pt}} +c}
\toprule
Method
& \multicolumn{2}{c}{Final dist $\downarrow$}
& \multicolumn{2}{c}{ADE $\downarrow$}
& \multicolumn{2}{c}{RPE $\downarrow$} \\
\midrule

Diffusion Policy
& $0.863$ &
& $0.4772$ &
& $0.0515$ & \\

\textbf{Particle Filter}
& $\mathbf{0.792}$ & $\pairedCI{ciGreen}{-11\%}{-5\%}$
& $\mathbf{0.4501}$ & $\pairedCI{ciGreen}{-8\%}{-4\%}$
& $\mathbf{0.0444}$ & $\pairedCI{ciGreen}{-15\%}{-13\%}$ \\

\gtrow PF w/ pose oracle
& $0.490$ & $\pairedCI{ciGreen}{-48\%}{-38\%}$
& $0.2750$ & $\pairedCI{ciGreen}{-46\%}{-39\%}$
& $0.0364$ & $\pairedCI{ciGreen}{-31\%}{-27\%}$ \\

\bottomrule
\end{tabular}%
}
\end{table}

\begin{table*}[t]
    \centering
    \caption{
      Particle filtering improves success rate (SR) and state estimates on challenging wrist-camera-only manipulation tasks.
    }
    \vspace{-0.5em}
    \label{tab:manip-results}
    \setlength{\tabcolsep}{4pt}
    \renewcommand{\arraystretch}{1.15}

    \resizebox{\textwidth}{!}{%
    \begin{tabular}{@{} l c c c c@{\hspace{0.18em}}c c@{\hspace{0.18em}}c c@{\hspace{0.18em}}c c@{\hspace{0.18em}}c @{}}
        \toprule
        & \multicolumn{3}{c}{Method Details}
        & \multicolumn{4}{c}{Wrist-Camera-Only Transport}
        & \multicolumn{4}{c}{Wrist-Camera-Only ToolHang} \\
        \cmidrule(lr){2-4}
        \cmidrule(lr){5-8}
        \cmidrule(lr){9-12}

        Method
        & Action selection
        & Num. particles
        & Persistent state?
        & \multicolumn{2}{c}{SR (\%) $\uparrow$}
        & \multicolumn{2}{c}{State RMSE $\downarrow$}
        & \multicolumn{2}{c}{SR (\%) $\uparrow$}
        & \multicolumn{2}{c}{State RMSE $\downarrow$} \\
        \midrule

        Diffusion Policy (DP)
        & -- & 1 &
        & $88.4$ &
        & $0.282$ &
        & $22.8$ &
        & $0.042$ & \\

        DP w/ persistent state
        & -- & 1 & $\checkmark$
        & $85.0$ & $\pairedCI{ciYellow}{-7.2}{+0.2}$
        & $0.324$ & $\pairedCI{ciRed}{+7\%}{+24\%}$
        & $\mathit{23.4}$ & $\pairedCI{ciYellow}{-4.2}{+5.4}$
        & $0.050$ & $\pairedCI{ciRed}{+7\%}{+28\%}$ \\\midrule

        DP ensemble
        & uniform mean & 32 &
        & $88.6$ & $\pairedCI{ciYellow}{-5.4}{+5.0}$
        & $0.216$ & $\pairedCI{ciGreen}{-29\%}{-19\%}$
        & $5.4$ & $\pairedCI{ciRed}{-22.0}{-13.2}$
        & $\mathit{0.032}$ & $\pairedCI{ciGreen}{-37\%}{-12\%}$ \\

        DP ensemble w/ persistent state
        & uniform mean & 32 & $\checkmark$
        & $\mathit{88.8}$ & $\pairedCI{ciYellow}{-4.8}{+5.0}$
        & $0.224$ & $\pairedCI{ciGreen}{-28\%}{-13\%}$
        & $8.8$ & $\pairedCI{ciRed}{-19.2}{-8.8}$
        & $0.034$ & $\pairedCI{ciGreen}{-38\%}{-2\%}$ \\\midrule

        PF w/ weighted action averaging
        & weighted mean & 32 & $\checkmark$
        & $\mathbf{92.0}$ & $\pairedCI{ciYellow}{-0.2}{+7.2}$
        & $\mathbf{0.187}$ & $\pairedCI{ciGreen}{-41\%}{-28\%}$
        & $11.2$ & $\pairedCI{ciRed}{-16.0}{-7.0}$
        & $\mathbf{0.031}$ & $\pairedCI{ciGreen}{-38\%}{-15\%}$ \\

        \textbf{PF w/ best-particle action selection}
        & best particle & 32 & $\checkmark$
        & $\mathbf{92.0}$ & $\pairedCI{ciGreen}{+0.4}{+7.0}$
        & $\mathit{0.194}$ & $\pairedCI{ciGreen}{-38\%}{-25\%}$
        & $\mathbf{28.6}$ & $\pairedCI{ciGreen}{+1.0}{+10.6}$
        & $0.033$ & $\pairedCI{ciGreen}{-30\%}{-15\%}$ \\

        \bottomrule
    \end{tabular}%
    }
\end{table*}

We additionally evaluate the 100,000-demonstration model on 100 out-of-distribution scenarios with eight random seeds (\cref{tab:ood-100k-comparison}).
While training scenes contain multiple randomly placed obstacles and randomly selected endpoints, OOD scenes contain a single rectangular or circular obstacle (dimensions and endpoints remain randomized).
Filtering improves both task performance and state estimation under this distribution shift, demonstrating that additional inference can help even where direct action generation performs better in-distribution.

\subsection{Extension to Visuomotor Manipulation}

\begin{figure}[t]
  \centering
  \begin{tikzpicture}
    \node[anchor=south west, inner sep=0] (img) {
      \includegraphics[width=0.925\linewidth]{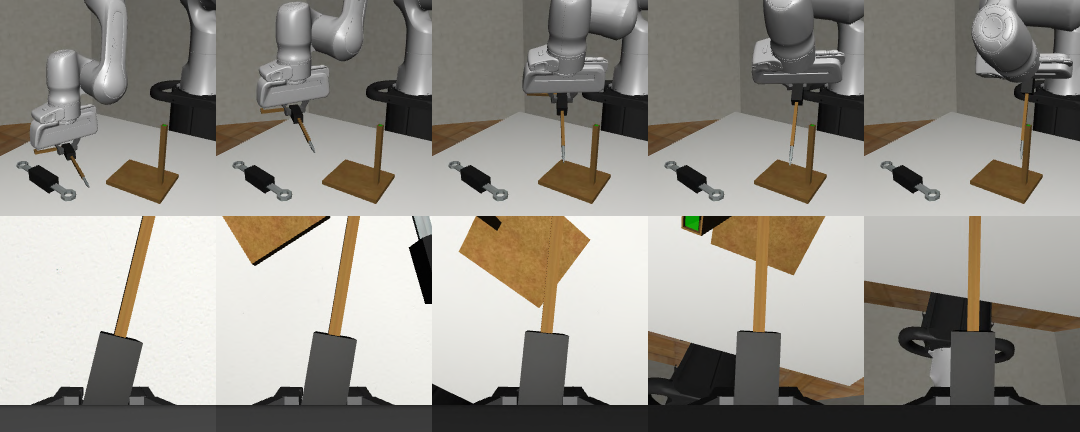}%
    };

    \path (img.south west) --
    node[pos=.25, xshift=-2.5mm, rotate=90] {
      \footnotesize wrist cam%
    }
    node[pos=.75, xshift=-2.5mm, rotate=90] {
      \footnotesize shoulder cam%
    }
    (img.north west);
  \end{tikzpicture}%
  \vspace{-0.75em}
  \caption{
    Camera observations from a validation set episode of the ToolHang task.
    \textbf{All of our manipulation experiments withhold access to the shoulder camera,} so the policy must infer actions and state solely from the wrist camera's view.}
  \label{fig:wrist-vs-shoulder-cam}
\end{figure}

\subsubsection{Datasets}

We consider two datasets based on the \textit{Transport} and \textit{ToolHang} tasks from Robomimic~\citep{mandlekar2021robomimic}.
States concatenate object positions and 6D rotations: three object poses for Transport (27 dimensions) and two for ToolHang (18 dimensions).
The original dataset includes two available camera views per manipulator: one fixed to the end effector (wrist camera) and another at a static external location (shoulder camera).
Since standard diffusion policies already perform strongly when all cameras are available, we instead remove the static cameras, keeping only the wrist camera.
As illustrated in \cref{fig:wrist-vs-shoulder-cam}, this makes the tasks substantially harder, since the shoulder camera shows task-relevant state in almost every individual frame, while the wrist camera sees only changing and partial views of the scene.

\subsubsection{Model \& Training}

The observation encoder~$\mathcal{E}_{\rm obs}$ uses the same convolutional image encoder architecture as Diffusion Policy~\citep{chi2023diffusionpolicy}.
A block-causal transformer is used for the denoising network, just like for the navigation task.

\subsubsection{Baselines}

For manipulation, we evaluate both stateless and persistent variants of Diffusion Policy and the ensemble baseline.
Stateless variants regenerate all state tokens within the current context window at each planning iteration, including those corresponding to past observations, conditioned on the observation-action history.
Persistent variants retain previously sampled states and feed them back as conditioning at subsequent iterations.
This comparison is possible because manipulation has no explicit goal condition to preserve, while in navigation, persistent states carry goal information forward after the initial goal token leaves the context window.
The persistent ensemble retains a separate state history for each member and uses uniform observation scores and uniform action averaging.
As in navigation, we compare best-particle action selection with weighted action averaging for particle filtering.

\subsubsection{Results (\texorpdfstring{\cref{tab:manip-results}}{Table \ref*{tab:manip-results}})}

We evaluate each method on 50 scenarios with 10 different random seeds, totaling 500 closed-loop simulations per task.
State estimates are computed as the componentwise weighted mean, and RMSE is computed across state components and timesteps.
State estimation errors are reported only for successful rollouts to reduce confounding from differences in task progress and failure modes.
We again report paired 95\% confidence intervals for changes from Diffusion Policy.

Using the same trained model, particle filtering with best-particle action selection improves success and reduces state RMSE relative to Diffusion Policy on both tasks.
These results extend the benefits of inference-time filtering to RGB-based manipulation under restricted visual observations.

Retaining sampled state histories does not automatically provide useful memory for control.
Single-particle persistence offers no clear success benefit, suggesting that carrying forward an uncorrected hypothesis can be insufficient under partial observability.
Maintaining multiple histories with uniform averaging likewise provides no reliable control improvement, even though it lowers state-estimation error.
This highlights the value of maintaining competing hypotheses and using observation evidence to determine their influence on control.

ToolHang highlights a potential limitation of averaging action proposals under ambiguity.
Proposals associated with different plausible behaviors may each be reasonable individually, yet yield ineffective actions when averaged.
Best-particle selection preserves an individual proposed action sequence and achieves better task performance despite slightly higher reported state RMSE.
The stronger dependence on the choice of action selection rule in ToolHang (as well as in our earlier navigation experiments) suggests that the consequences of action aggregation vary across tasks.

\section{Conclusion}

We presented an inference-time approach that uses a joint generative model of states, actions, and observations for both particle filtering and control.
Each particle maintains a state-history hypothesis and an associated action proposal, while agreement between predicted and measured observations updates its weight.
These weights guide action selection, allowing observation evidence to influence control through persistent hypotheses.
The procedure uses the same trained model as direct action generation, without additional filtering-specific training or separately learned dynamics, observation, or policy models.

Experiments in partially observed navigation and manipulation demonstrate that this approach can improve both state estimation and task performance.
Controlled studies demonstrate the importance of useful state-action hypotheses, observation-consistent weighting, particle coverage, and the choice of action-selection rule.
In navigation, filtering improves control with limited training data and remains useful even with a large synthetic training dataset when obstacle layouts differ from those seen during training.
For manipulation, where collecting demonstrations is costly, filtering improves control without requiring more training data.

Particle filtering also provides a natural way to incorporate information available at deployment, such as known maps or explicit sensor models, by using it to assess generated hypotheses without retraining the model.
More broadly, it provides a principled framework for using additional inference-time computation with pretrained joint trajectory models.

\balance
\bibliographystyle{IEEEtranN}
\bibliography{main}  %

\end{document}